\documentclass[runningheads]{llncs}
\usepackage{graphicx}
\usepackage[T5]{fontenc}
\usepackage[utf8]{inputenc}
\usepackage{cite}
\usepackage{amssymb}
\usepackage{amsmath}
\usepackage{multirow}
\usepackage{url}
\usepackage{tabularx}
\usepackage{pgfplots}
\usepackage{float}
\usepackage{sidecap}
\usepackage{wrapfig}
\usepackage{hyperref}
\usepackage{makecell}
\usepackage{appendix}
\usepackage{longtable}
\usepackage{caption}

\usetikzlibrary{patterns}

\begin{document}
\title{A Large-scale Dataset for Hate Speech Detection on Vietnamese Social Media Texts}

\author{Son T. Luu\inst{1,2} \and
Kiet Van Nguyen\inst{1,2} \and
Ngan Luu-Thuy Nguyen\inst{1,2}}

\institute{University of Information Technology, Ho Chi Minh City, Vietnam \and
Vietnam National University Ho Chi Minh City, Vietnam \\
\email{\{sonlt,kietnv,ngannlt\}@uit.edu.vn}}

\newcommand{\footref}[1]{%
    $^{\ref{#1}}$%
}
\newcolumntype{L}[1]{>{\raggedright\let\newline\\\arraybackslash\hspace{0pt}}m{#1}}
\newcolumntype{C}[1]{>{\centering\let\newline\\\arraybackslash\hspace{0pt}}m{#1}}
\newcolumntype{R}[1]{>{\raggedleft\let\newline\\\arraybackslash\hspace{0pt}}m{#1}}

\titlerunning{ }
\authorrunning{ } 

\maketitle  

\begin{abstract}
In recent years, Vietnam witnesses the mass development of social network users on different social platforms such as Facebook, Youtube, Instagram, and Tiktok. On social media, hate speech has become a critical problem for social network users. To solve this problem, we introduce the ViHSD - a human-annotated dataset for automatically detecting hate speech on the social network. This dataset contains over 30,000 comments, each comment in the dataset has one of three labels: CLEAN, OFFENSIVE, or HATE. Besides, we introduce the data creation process for annotating and evaluating the quality of the dataset. Finally, we evaluate the dataset by deep learning and transformer models. 

\end{abstract}

\section{Introduction}
\label{introduction}
There are approximately 70 million Internet users in Vietnam, and most of them are familiar with Facebook and Youtube. Nearly 70\% of total Vietnamese people use Facebook, and spend an average of 2.5 hours per day on it \cite{NhaNguyen2020Gender}. On social network, hate easily appears and spread. According to Kang and Hall \cite{kang2020hate}, hate-speech does not reveal itself as hatred. It is just a mechanism to protect an individual's identity and realm from the others. Hate leads to the destruction of humanity, isolating people, and debilitating society. Within the development of social network sites, the hate appears on social media as hate-speech comments, hate-speech posts, or messages, and it spreads too fast. The existence of hate speech makes the social networking spaces toxic, threatens social network users, and bewilders the community.

The automated hate speech detection task is categorized as the supervised learning task, specifically closed to the sentiment analysis task\cite{schmidt2017survey}. There are several state-of-the-art approaches such as deep learning and transformer models for sentiment analysis. However, to be able to make experiment on hate speech detection task, the datasets, especially large-scale datasets, play an important role. To handle this, we introduce ViHSD - a large-scale dataset used for automatically hate speech detection on Vietnamese social media texts to overcome the hate-speech problem on social networks. Then, we present the annotation process for our dataset and the method to ensure the quality of annotators. Finally, we evaluate our dataset on SOTA models and analyze the obtained empirical results to explore the advantages and disadvantages of the models on the dataset.

The content of the paper is structured as follows. Section \ref{related_work} takes an overview on current researching for hate-speech detection. Section \ref{dataset} shows statistical figures about our dataset as well as our annotation procedure. Section \ref{methodologies} presents the classification models applied for our dataset to solve the hate-speech detection problem. Section \ref{empirical} describes our experiments on the dataset and the analytical results. Finally, Section \ref{conclusion} concludes and proposes future works.
\section{Related Works}
\label{related_work}
In English as well as other languages, there are many datasets constructed for hate speech detection. We divided them into two categories: flat labels and hierarchical labels. According to Cerri et al. \cite{CERRI201439}, flat labels are treated as no relation between different labels. In contrast, hierarchical labels has a hierarchical structure which one or more labels can have sub-labels, or being grouped in super-labels. Besides, Hatebase \cite{tuckwood2017hatebase} and Hurtlex \cite{Bassignana2018HurtlexAM} are two abusive words sets used for lexicon-based approaching for the hate speech detection problem. 

For flat labels, we introduce two typical and large-scale datasets in English. The first dataset is provided by Waesem and Hovy \cite{waseem-hovy-2016-hateful}, which contains 17,000 tweets from Twitter and has three labels: racism, sexism, and none. The second dataset is provided by Davidson et al. \cite{Davidson2017AutomatedHS}, which contains 25,000 tweets from Twitter and also has three labels including: hate, offensive, and neither. Apart from English, there are other datasets in other languages such as: Arabic \cite{8508247}, and Indonesian \cite{8355039}.

For hierarchical labels datasets, Zampieri et al. \cite{zampieri-etal-2019-predicting} provide a multi-labelled dataset for predicting offensive posts on social media in English. This dataset serves two tasks: group-directed attacking and person-directed attacking, in which each task has binary labels. Another similar multi-labelled dataset in English are provided by Basile et al. \cite{basile-etal-2019-semeval} in SemEval Task 5 (2019). Other multi-labelled datasets in other non-English languages are also constructed and available such as:  Portuguese \cite{fortuna-etal-2019-hierarchically}, Spanish \cite{fersini2018overview}, and Indonesian \cite{ibrohim-budi-2019-multi}. Besides, multilingual hate speech corpora are also constructed such as hatEval with English and Spanish \cite{basile-etal-2019-semeval} and CONAN with English, French, and Italian \cite{chung-etal-2019-conan}. 

The VLSP-HSD dataset provided by Vu et al. \cite{sonvx2019} is a dataset used for the VLSP 2019 shared task about Hate speech detection on Vietnamese language\footnote{\url{https://www.aivivn.com/contests/8}}. However, the authors did not mention the annotation process and the method for evaluating the quality of the dataset. Besides, on the hate speech detection problem, many state-of-the-art models give optimistic results such as deep learning models \cite{10.1145/3041021.3054223} and transformer language models \cite{isaksen-gamback-2020-using}. Those models require large-scale annotated datasets, which is a challenge for low-resource languages like Vietnamese. Moreover, current researches about hate speech detection do not focus on analyzing about the sentiment aspect of Vietnamese hate speech language. Those ones are our motivation to create a new dataset called ViHSD for Vietnamese with strict annotation guidelines and evaluation process to measure inter-annotator agreement between annotators.

\section{Dataset Creation}
\label{dataset}

\subsection{Data Preparation}
We collect users' comments about entertainment, celebrities, social issues, and politics from different Vietnamese Facebook pages and YouTube videos. We select Facebook pages and YouTube channels that have a high-interactive rate, and do not restrict comments. After collected data, we remove the name entities from the comments in order to maintain the anonymity.

\subsection{Annotation Guidelines}
The ViHSD dataset contains three labels: HATE, OFFENSIVE, and CLEAN. Each annotator assigns one label for each comment in the dataset. In the ViHSD dataset, we have two labels denoting for hate speech comments, and one label denoting for normal comments. The detailed meanings about three labels and examples for each label are described in Table \ref{tab_annotation_guideline}. 

Practically, many comments in the dataset are written in informal form. Comments often contain abbreviation such as \textit{M.n} (English: Everyone), \textit{mik} (English: us) in \textbf{Comment 1} and \textit{Dm} (English: f*ck) in \textbf{Comment 2}, and slangs such as \textit{chịch} (English: f*ck), \textit{cái lol} (English: p*ssy) in \textbf{Comment 2}. Besides, comments has the figurative meaning instead of explicit meaning. For example, the word: \textit{lũ quan ngại} (English: dummy pessimists) in \textbf{Comment 3} is usually used by many Vietnamese Facebook users on social media platform to mention a group of people who always think pessimistically and posting negative contents.

\begin{table}[ht!]
    \caption{Annotation guidelines for annotating Vietnamese comments in the Hate Speech Detection task. }
    \label{tab_annotation_guideline}
    \resizebox{\textwidth}{!}{
    \begin{tabular}{|C{2cm}|p{5.1cm}|p{6.4cm}|}
        \hline
        \textbf{Label} & \makecell{\centering \textbf{Description}} & \makecell{\centering \textbf{Example}} \\
        \hline
        CLEAN & \makecell*[{{p{5cm}}}]{The comments have no harassment at all.} & \makecell*[{{p{6.3cm}}}]{\textbf{Comment 1}: M.n ơi cho mik hỏi mik theo dõi cô mà mik hk pít cô là con gái thiệt hả m.n (\textit{English}: Hey everyone! Is she a truly girl?) \\ (This comment is thoroughly clean, in which there are no bad words or profane language, and does not attack anyone) } \\
        \hline
        OFFENSIVE & \makecell*[{{p{5cm}}}]{The comments contain harassment contents, even profanity words, but do not attack any specific object.} & \makecell*[{{p{6.3cm}}}]{ \textbf{Comment 2}: Đồ  \underline{khùng} (\textit{English}: Madness) \\ (This comment has offensive word \textit{"khùng"}. Nevertheless, it does not contain any word that aims to a person or a group. In addition, \textit{"khùng"} is also a slang, which means mad) } \\
        \hline
        HATE & \makecell*[{{p{5cm}}}]{The comments have harassment and abusive contents and directly aim at an individual or a group of people based on their characteristics, religion, and nationality. \\ Some exceptional cases happened with the HATE label: \\ \textbf{Case 1}: The comments have offensive words and attack a specific object such as an individual, a community, a nation, or a religion. This case is easy to identify hate speech. \\ \textbf{Case 2}: The comments have racism, harassment, and hateful meaning, however, does not contain explicit words. \\ \textbf{Case 3}: The comments have racism, harassment, and hateful meaning, but showed as figurative meaning. To identify this comment, users need to have particular knowledge about social. } & \makecell*[{{p{6.3cm}}}]{\textbf{Comment 3}: Dành cho \underline{lũ quan ngại} (\textit{English: This is for those dummy pessimists}) \\ (This comment contains a phrase, which is underlined, mentions to a group of people with bad meaning) \\ \textbf{Comment 4}: \underline{Dm} Có a mới không ổn. Mày rình mày \underline{chịch} riết ổn \underline{cái lol} (\textit{English: F*ck you. I am not fine. You are fine why you're making sex ?}) \\ (This comments contained many of profanities, which are underlined. Besides, it contains personal pronoun "Mày", which aims to a specific person) \\ \textbf{Comment 3}: Ở đấy ngột ngạt quá thì đưa nó qua <LOC> cho nó thoáng mát (\textit{English: If this place is so stifling and not suitable for bitch like you, why don't you choose <LOC>? }) \\ (This comment has the phrase <LOC> mentioned to a specific location, which has racism meaning. However, this comment does not has any bad words at all)} \\
        \hline
    \end{tabular}
    }
\end{table}

\subsection{Data Creation Process}
\begin{figure}[h]
\centering
    \includegraphics[scale=0.36]{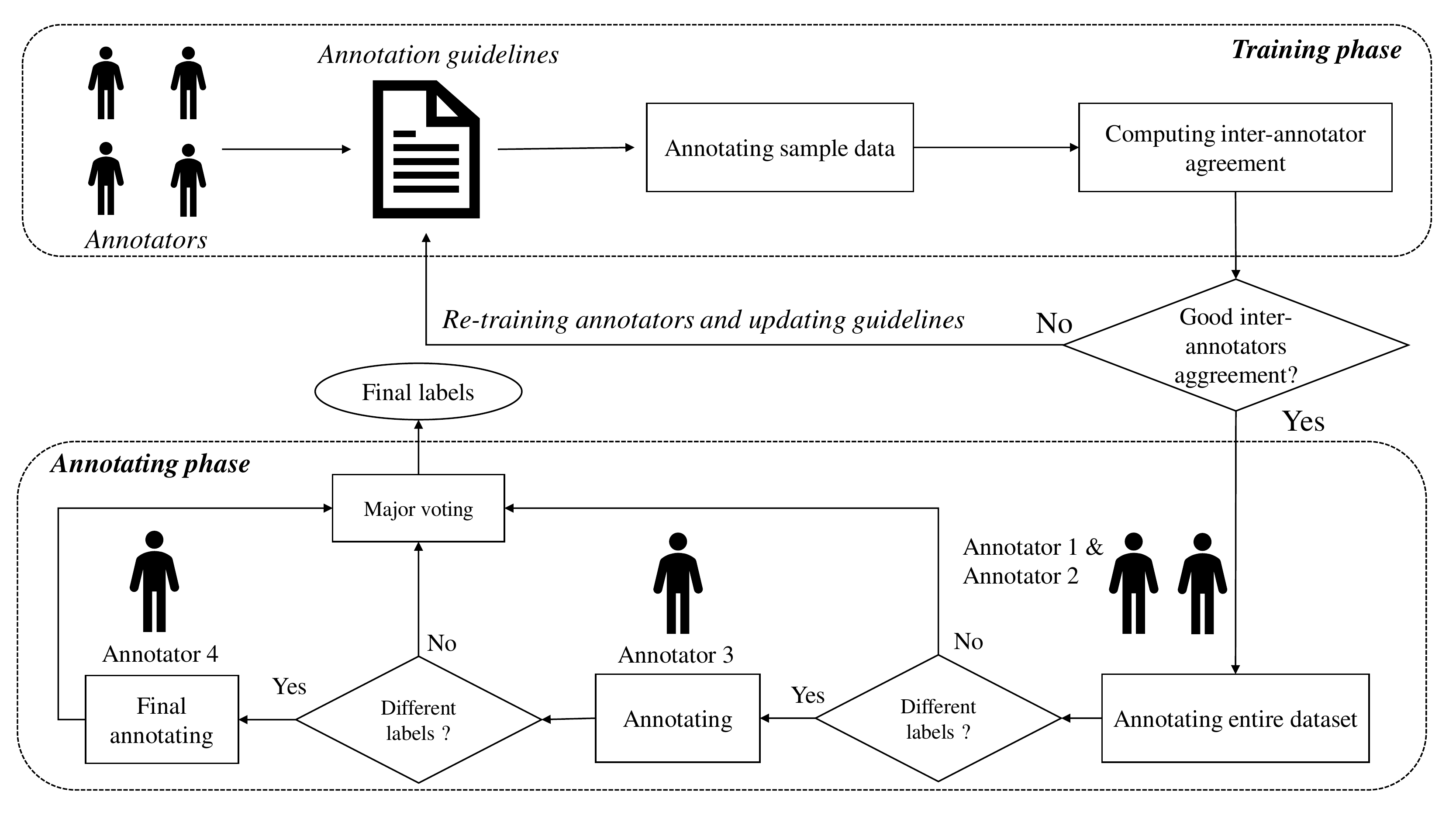}
    \caption{Data annotation process for the ViHSD dataset. Acccording to Eugenio\cite{di-eugenio-2000-usage}, $\kappa > 0.5$ is acceptable. }
    \label{fig_annotation_process}
\end{figure}

Our annotation process contains two main phases as described in Figure. \ref{fig_annotation_process}. The first one is the training phase, which annotators are given a detailed guidelines, and annotate for a sample of data after reading carefully. Then we compute the inter-annotator agreement by Cohen Kappa index ($\kappa$) \cite{doi:10.1177/001316446002000104}. If the inter-annotators agreement not good enough, we will re-train the annotator, and re-update the annotation guidelines if necessary. After all annotators are well-trained, we go to annotation phase. Our annotation phase is inspired from the IEEE peer review process of articles \cite{7902269}. Two annotators annotate the entire dataset. If there are any different labels between two annotators, we let the third annotators annotate those labels. The fourth annotators annotate if all three annotators are disagreed. The final label are defined by Major voting. By this way, we guaranteed that each comment is annotated by one label and the objectivity for each comment. Therefore, the total time spent on annotating is less than four annotators doing with the same time. 

\subsection{Dataset Evalutation and Discussion}
We randomly take 202 comments from the dataset and give them to four different annotators, denoted as A1, A2, A3 and A4, for annotating. Table \ref{tbl_inter_annotator_agreement} shows the inter-annotator agreement between each pair of annotators. Then, we compute the average inter-annotator agreement. The final inter-annotator agreement for the dataset is $\kappa=0.52$.

\begin{table}[H]
    \centering
    \caption{The confusion matrix between annotators in a set of 202 comments computed by Cohen Kappa index ($\kappa$).}
    \label{tbl_inter_annotator_agreement}
    \begin{tabular}{|c|c|c|c|c|}
        \hline
        & A1 & A2 & A3 & A4 \\
        \hline
        A1 & - & 0.46 & 0.51 & 0.65\\
        \hline
        A2 & - & - & 0.47 & 0.53 \\
        \hline
        A3 & - & - & - & 0.55 \\
        \hline
        A4 & - & - & - & - \\
        \hline
    \end{tabular}
    
\end{table}

The ViHSD dataset was crawled from the social network so they had many abbreviations, informal words, slangs, and figurative meaning. Therefore, it confuses annotators. For example, the \textbf{Comment 1} contains the phrase: \textit{mik}, which mean \text{mình (English: I)}, and the \textbf{Comment 4} in Table \ref{tab_annotation_guideline} has the profane word \textit{Dm} (English: m*ther f**ker). Assume that two annotators assign label for \textbf{Comment 4}, and it contains the word \textit{Dm} written in abbreviation form. The first annotator knew about this word before, thus he/she annotates this comment as hate. The second annotator instead, annotates this comment as clean because he/she do not understand that word. The next example is the phrase \textit{lũ quan ngại} (English: dummy pessimists) in the \textbf{Comment 3} in Table \ref{tab_annotation_guideline}. Two annotators assign label for \textbf{Comment 3}. The first annotator does not understand the real meaning of that phase, thus he/she marks this comment as clean. In contrast, the second annotator knows what is the real meaning of this word (See Section 3.2 for the meaning of that phase) so he/she knows the abusive meaning of \textbf{Comment 3} and annotates it as hate. Although the guidelines have clearly definition about the CLEAN, OFFENSIVE, and HATE labels, the annotation process is mostly impacted by the knowledge and subjective of annotators. Thus, it is necessary to re-train annotators and improve the guidelines continuously to increase the quality of annotators and the inter-annotator agreement.

\subsection{Dataset Overview}
\begin{table}[H]
    \centering
    \caption{Several examples extracted from the ViHSD dataset.}
        \label{tbl_dataset_examples}
           \resizebox{0.97\textwidth}{!}{
                \begin{tabular}{|C{0.5cm}|p{11.1cm}|C{1.5cm}|}
                \hline
                \textbf{\#} & \makecell{\centering \textbf{Comments}} & \textbf{Label} \\
                \hline
                1 & \makecell*[{{p{11cm}}}]{Nhanh thực sự  (\textit{English: It is really fast} )} & clean \\
                \hline
                2 & \makecell*[{{p{11cm}}}]{Đm chứ biết làm gì (\textit{English: How f*ck damn to do that!} ) } & offensive \\
                \hline
                3 & \makecell*[{{p{11cm}}}]{Nó học cách của thằng anh nó đó, hèn, khốn nạn (\textit{English: He is coward and bastard likes his brother} )} & hate \\
                \hline
                4 & \makecell*[{{p{11cm}}}]{<person name> người ư? sinh vật hạ đẳng chứ ngươi ai lam thế (\textit{English: <person name> person ? It is a inferior animal. Human don't do that} ) } & hate \\
                \hline
                5 & \makecell*[{{p{11cm}}}]{Đm vcl (\textit{English: God damn it!!} ) } & offensive \\
                \hline
                \end{tabular}
        }
\end{table}

\begin{figure}[H]
    \centering
    \begin{tikzpicture}[scale=0.85]
            \begin{axis}[
                ybar,
                enlarge y limits={0.45,upper},
                enlarge x limits=0.25,
                symbolic x coords={TRAIN, DEV, TEST},
                xtick=data,
                ymin = 0, ymax = 19000,
                nodes near coords,
                every node near coord/.append style={rotate=90, anchor=west},
        	    ylabel near ticks,
        	    ylabel={Numnber of comments},
        	    x tick label
        	   style={font=\footnotesize}
            ]
            \addplot[draw=black, pattern=north east lines] coordinates {
                (TRAIN, 19886) 
                (DEV,2190) 
                (TEST,5548)
            };
            \addplot[draw=black, pattern=dots] coordinates {
                (TRAIN, 1606) 
                (DEV,212) 
                (TEST,444)
            };
            \addplot[draw=black, fill=gray] coordinates {
                (TRAIN, 2556) 
                (DEV,270) 
                (TEST,688)
            };
            \legend{CLEAN, OFFENSIVE, HATE}
            \end{axis}
    \end{tikzpicture}
    \caption{The distributions of three labels on the train, dev, and test sets.}
    \label{fig:data_distribution_train_set}
\end{figure}
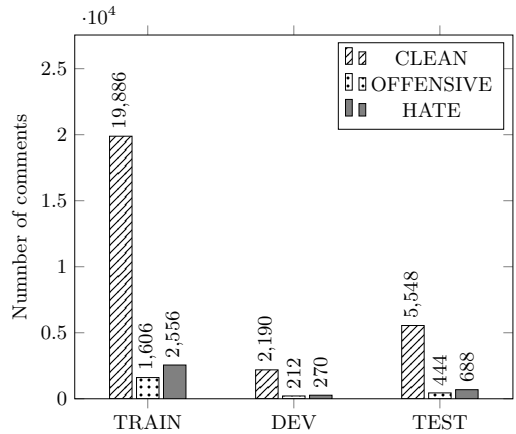

The ViHSD contains 33,400 comments. Each one was labelled as CLEAN (0), OFFENSIVE (1), and HATE (2). Table \ref{tbl_dataset_examples} displays some examples from the ViHSD dataset. Then we divided our dataset into training (train), development (dev), and test sets, respectively, with proportion: 7-1-2. Fig \ref{fig:data_distribution_train_set} describes the distribution of data on three labels on those sets. According to Fig \ref{fig:data_distribution_train_set}, the distribution of data labels on the training, development, and test sets are the same, and the data are skewed to the CLEAN label. 
\section{Baseline Models}
\label{methodologies}

The problem of text classification is defined according to Aggarwal and Zhai \cite{Aggarwal2012} as given a set of training texts as training data $D=\{X_1, X_2, ..., X_n\}$, in which each $X_i \in D$ has one of the label in the label set $\{1..k\}$. The training data is used to build the classification model. Then, for unlabeled data coming, the classification model predicts the label for it. In this section, we introduce two approaches for constructing prediction models on the ViHSD dataset.

\subsection{Deep Neural Network Models (DNN Models)}
Convolutional Neural Network (CNN) uses particular layers called the CONV layer for extracting local features in the image \cite{726791}. However, although invented for computer vision, CNN can also be applied to Natural Language Processing (NLP), in which a filter W relevant to a window of $h$ words \cite{kim-2014-convolutional}. Besides, the pre-trained word vectors also influence the performance of the CNN model \cite{kim-2014-convolutional}. Besides, Gated Recurrent Unit (GRU) is a variant of the RNN network. It contains two recurrent networks: encoding sequences of texts into a fixed-length word vector representation, and another for decoding word vector representation back to raw input \cite{cho-etal-2014-learning}. 

We implement the Text-CNN and the GRU models, and evaluate them on the ViHSD dataset with the \textit{fasttext\footnote{ \url{https://fasttext.cc/docs/en/crawl-vectors.html}}} pre-trained word embedding of 157 different languages provided by Grave et al. \cite{grave-etal-2018-learning}. This embedding transforms a word into a 300 dimension vector. 

\subsection{Transformer Models}
The transformer model \cite{vaswani2017attention} is a deep neural network architecture based entirely on the attention mechanism, replaced the recurrent layers in auto encoder-decoder architectures with special called multi-head self-attention layers. Yang et al.  \cite{yang2019comprehensive} found that the transformer blocks improved the performance of the classification model. In this paper, we implement BERT\cite{devlin2019} - the SOTA transformer model with multilingual pre-training\footnote{\url{https://github.com/google-research/bert/blob/master/multilingual.md}} such as bert-base-multilingual-uncased (m-BERT uncased) and bert-base-multilingual-cased (m-BERT cased), DistilBERT \cite{sanh2020distilbert} - a lighter but faster variant of BERT model with multilingual cased pre-trained model\footnote{\url{https://huggingface.co/distilbert-base-multilingual-cased}}, and XLM-R\cite{conneau-etal-2020-unsupervised} - a cross-lingual language model with xlm-roberta-base pre-trained\footnote{\url{https://huggingface.co/xlm-roberta-base}} . Those multilingual pre-trained models are trained on various of languages including Vietnamese. 

\section{Experiment}
\label{empirical}
\subsection{Experiment Settings}
First of all, we pre-process our dataset as belows: (1) Word-segmentating texts into words by the pyvi tool\footnote{\url{https://pypi.org/project/pyvi/}}, (2) Removing stopwords\footnote{List Vienamese stopwords \url{https://github.com/stopwords/vietnamese-stopwords}}, (3) Changing all texts into lower cases\footnote{We do not lower case texts with cased pre-trained transformer models}, and (4) Removing special characters such as hashtags, urls, and mention tags.

Next, we run the Text-CNN model with 50 epochs, batch size equal to 256, sequence length equal to 100, and the dropout ratio is 0.5. Our model uses 2D Convolution Layer with 32 filters and size 2, 3, 5, respectively. Then, we run the GRU model with 50 epochs, sequence length equal to 100, the dropout ratio is 0.5, and the bidirectional GRU layer. We use the Adam optimizer for both Text-CNN and GRU. Finally, we implement transformer models includes BERT, XLM-R, and DistilBERT with the batch size equal to 16 for both training and evaluation, 4 epochs, sequence length equal to 100, and manual seed equal to 4. 

\subsection{Experiment Results}

Table \ref{tab_empirical_result} illustrates the results of deep neural models and transformer models on the ViHSD dataset. The results are measured by Accuracy and macro-averaged F1-score. According to Table \ref{tab_empirical_result}, Text-CNN achieves 86.69\% in accuracy and 61.11\% in F1-score, which is better than GRU. Transformer models such as BERT, XLM-R, and DistilBERT give better results than deep neural models in F1-score. The BERT with bert-base-multilingual-cased model (m-bert cased) obtained best result in both Accuracy and F1-score with 86.88\% and 62.69\% respectively on the ViHSD dataset.

\begin{table}[ht!]
    \caption{Empirical results of baseline models on the test set}
    \label{tab_empirical_result} 
    \begin{tabular}{|p{1.9cm}|p{1.8cm}|p{4.8cm}|R{1.6cm}|R{1.6cm}|}
        \hline
        & \textbf{Model} & \textbf{\makecell[c]{Pre-trained model}} & \textbf{\makecell[tl]{Accuracy \\ (\%)}} & \textbf{\makecell[tl]{F1-macro \\ (\%)}} \\
        \hline
        \multirow{2}{1.8cm}{DNN Models} & \makecell[lt]{Text CNN} & fastText & 86.69 & 61.11 \\
        \cline{2-5}
        & GRU & fastText & 85.41 & 60.47 \\
        \hline
        \multirow{4}{1.8cm}{Transformer models} & \multirow{2}{*}{BERT} & bert-base-multilingual-uncased & 86.60 & 62.38 \\
        \cline{3-5}
        & & bert-base-multilingual-cased & \textbf{86.88} & \textbf{62.69} \\
        \cline{2-5}
        & \multirow{1}{*}{XLM-R} & xlm-roberta-based & 86.12 & 61.28 \\
        \cline{2-5}
        & \multirow{1}{*}{DistilBERT} & distilbert-base-multilingual-cased & 86.22 & 62.42 \\
        \hline
        
    \end{tabular}
    
\end{table}

Overall, the performance of transformer models are better than deep neural models, indicating the power of BERT and its variants on text classification task, especially on hate speech detection even if they were trained on various languages. Additionally, there is a large gap between the accuracy score and the F1-score, which caused by the imbalance in the dataset, as described in Section \ref{dataset}

\subsection{Error Analysis}
\begin{figure}[H]
\centering
    \includegraphics[scale=0.55]{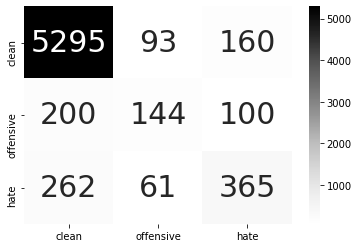}
    \caption{Confusion matrix of the m-BERT cased model on the ViHSD dataset}
    \label{fig_cf_textcnn}
\end{figure}

Figure. \ref{fig_cf_textcnn} shows the confusion matrix of the m-BERT cased model. Most of the offensive comments in the dataset are predicted as clean comments. Besides, Table \ref{tab_error_sample} shows incorrect predictions by the m-BERT cased model. The comments number 1, 2, and 3 had many special words, which are only used on the social network such as: \textit{"dell"}, \textit{"coin card"}, and \textit{"éo"}. These special words make those comments had wrong predicting labels, misclassified from offensive labels to clean labels. Moreover, the comments number 4 and 5 have profane words and are written in abbreviation form and teen codes such as \textit{"cc"}, \textit{"lol"}. Specifically, for the fifth comments, in which \textit{"3"} represents for \textit{"father"} in English, combined which other bad words such as: \textit{"Cc"} - profane word and \textit{"m"} - represents for \textit{"you"} in English. Generally, the fifth comment has bad meaning (see Table \ref{tab_error_sample} for the English meaning) thus it is the hate speech comment. However, the classification model predicts that comment as an offensive label because it cannot identify the target objects and the profane words written in irregular form.

\begin{table}
    \caption{Wrong prediction samples in the ViHSD dataset}
    \label{tab_error_sample}
    \resizebox{\textwidth}{!}{
    \begin{tabular}{|C{0.5cm}|L{8.8cm}|R{0.9cm}|R{1.5cm}|}
        \hline
        \# & \makecell[c]{\textbf{Comments}} & \textbf{True} & \textbf{Predict} \\
        \hline
        1 & \makecell*[{{p{8.7cm}}}]{coin card :3 (\textit{English: F*ck d*ck }) } & 1 & 0 \\
        \hline
        2 & \makecell*[{{p{8.7cm}}}]{<person name> dell hiểu kiểu gì :)) (\textit{English: I do not f*cking understand why}) } & 1 & 0 \\
        \hline
        3 & \makecell*[{{p{8.7cm}}}]{Éo thích thì biến đi thọc con mắt chóng cái tai lên xem làm ghì (\textit{English: Get out, damn! Prick your eyes if you don't care.}) } & 1 & 0 \\
        \hline
        4 & \makecell*[{{p{8.7cm}}}]{<person name>chửi cc  (\textit{English: f*ck you}) } & 2 & 1 \\
        \hline
        5 & \makecell*[{{p{8.7cm}}}]{Cc 3 m (\textit{English: Like your dad's d*ck}) } & 2 & 1 \\
        \hline
    \end{tabular}
    }
\end{table}

In general, the wrong prediction samples and most of the ViHSD dataset comments are written with irregular words, abbreviations, slangs, and teencodes. Therefore, in the pre-processing process, we need to handle those characteristic to enhance the performance of classification models. For abbreviations and slangs, we can build a Vietnamese slangs dictionary for slangs replacement and a Vietnamese abbreviation dictionary to replace the abbreviations in the comments. Besides, for words that are not found in mainstream media, we can try to normalize them to the regular words. For example, words like \textit{gìiiiii}, \textit{ừmmmm} should be normalized to \textit{gì} (what) and \textit{ừ} (ok). In addition, the emoji icons are also a polarity feature to define whether a comment is negative or positive, which can support for detecting the hate and offensive content in the comments. 

\section{Conclusion}
\label{conclusion}
We constructed a large-scale dataset called the ViHSD dataset for hate speech detection on Vietnamese social media texts. The dataset contains 33,400 comments annotated by humans and achieves 62.69\% by the Macro F1-score with the BERT model. We also proposed an annotation process to save time for data annotating. 

The current inter-annotator agreement of the ViHSD dataset is just in moderate level. Therefore, our next studies focus on improving the quality of the dataset based on the data annotation process. Besides, the best baseline model on the dataset is 62.69\%, and this is a challenge for future researches to improve the performance of classification models for Vietnamese hate-speech detection task. From the error analysis, we found that it is difficult to detect the hate speech on Vietnamese social media texts due to their characteristics. Hence, we will improve the pre-processing technique for social media texts such as lexicon-based approach for teen codes and normalizing acronyms on next studies to increase the performance of this task.


\bibliographystyle{splncs04}
\bibliography{reference}

\begin{thebibliography}{10}
\providecommand{\url}[1]{\texttt{#1}}
\providecommand{\urlprefix}{URL }
\providecommand{\doi}[1]{https://doi.org/#1}

\bibitem{Aggarwal2012}
Aggarwal, C.C., Zhai, C.: A Survey of Text Classification Algorithms. Springer
  US, Boston, MA (2012)

\bibitem{8508247}
{Albadi}, N., {Kurdi}, M., {Mishra}, S.: Are they our brothers? analysis and
  detection of religious hate speech in the arabic twittersphere. In: IEEE/ACM
  International Conference on Advances in Social Networks Analysis and Mining
  (ASONAM). pp. 69--76 (2018)

\bibitem{8355039}
{Alfina}, I., {Mulia}, R., {Fanany}, M.I., {Ekanata}, Y.: Hate speech detection
  in the indonesian language: A dataset and preliminary study. In: 2017
  International Conference on Advanced Computer Science and Information Systems
  (ICACSIS). pp. 233--238 (2017)

\bibitem{10.1145/3041021.3054223}
Badjatiya, P., Gupta, S., Gupta, M., Varma, V.: Deep learning for hate speech
  detection in tweets. p. 759–760. WWW '17 Companion, International World
  Wide Web Conferences Steering Committee, Republic and Canton of Geneva, CHE
  (2017)

\bibitem{basile-etal-2019-semeval}
Basile, V., Bosco, C., Fersini, E., Nozza, D., Patti, V., Rangel~Pardo, F.M.,
  Rosso, P., Sanguinetti, M.: {S}em{E}val-2019 task 5: Multilingual detection
  of hate speech against immigrants and women in twitter. In: Proceedings of
  the 13th International Workshop on Semantic Evaluation. Association for
  Computational Linguistics, Minneapolis, Minnesota, USA

\bibitem{Bassignana2018HurtlexAM}
Bassignana, E., Basile, V., Patti, V.: Hurtlex: A multilingual lexicon of words
  to hurt. In: CLiC-it (2018)

\bibitem{CERRI201439}
Cerri, R., Barros, R.C., {de Carvalho}, A.C.: Hierarchical multi-label
  classification using local neural networks. Journal of Computer and System
  Sciences  \textbf{80}(1),  39 -- 56 (2014)

\bibitem{cho-etal-2014-learning}
Cho, K., van Merri{\"e}nboer, B., Gulcehre, C., Bahdanau, D., Bougares, F.,
  Schwenk, H., Bengio, Y.: Learning phrase representations using {RNN}
  encoder{--}decoder for statistical machine translation. In: Proceedings of
  the 2014 Conference on Empirical Methods in Natural Language Processing
  ({EMNLP}). pp. 1724--1734. Association for Computational Linguistics, Doha,
  Qatar (2014)

\bibitem{chung-etal-2019-conan}
Chung, Y.L., Kuzmenko, E., Tekiroglu, S.S., Guerini, M.: {CONAN} - {CO}unter
  {NA}rratives through nichesourcing: a multilingual dataset of responses to
  fight online hate speech. In: Proceedings of the 57th Annual Meeting of the
  Association for Computational Linguistics. Association for Computational
  Linguistics, Florence, Italy (2019)

\bibitem{doi:10.1177/001316446002000104}
Cohen, J.: A coefficient of agreement for nominal scales. Educational and
  Psychological Measurement  \textbf{20}(1),  37--46 (1960)

\bibitem{conneau-etal-2020-unsupervised}
Conneau, A., Khandelwal, K., Goyal, N., Chaudhary, V., Wenzek, G., Guzm{\'a}n,
  F., Grave, E., Ott, M., Zettlemoyer, L., Stoyanov, V.: Unsupervised
  cross-lingual representation learning at scale. In: Proceedings of the 58th
  Annual Meeting of the Association for Computational Linguistics. pp.
  8440--8451. Association for Computational Linguistics, Online (2020)

\bibitem{Davidson2017AutomatedHS}
Davidson, T., Warmsley, D., Macy, M., Weber, I.: Automated hate speech
  detection and the problem of offensive language (2017)

\bibitem{devlin2019}
Devlin, J., Chang, M.W., Lee, K., Toutanova, K.: {BERT}: Pre-training of deep
  bidirectional transformers for language understanding. In: Proceedings of the
  2019 Conference of the North {A}merican Chapter of the Association for
  Computational Linguistics: Human Language Technologies, Volume 1 (Long and
  Short Papers). Association for Computational Linguistics, Minneapolis,
  Minnesota (2019)

\bibitem{di-eugenio-2000-usage}
Di~Eugenio, B.: On the usage of kappa to evaluate agreement on coding tasks.
  In: Proceedings of the Second International Conference on Language Resources
  and Evaluation ({LREC}{'}00). European Language Resources Association (ELRA),
  Athens, Greece (May 2000)

\bibitem{7902269}
{El-Hawary}, M.E.: What happens after i submit an article? [editorial]. IEEE
  Systems, Man, and Cybernetics Magazine  \textbf{3}(2),  3--42 (2017)

\bibitem{fersini2018overview}
Fersini, E., Rosso, P., Anzovino, M.: Overview of the task on automatic
  misogyny identification at ibereval 2018. IberEval@ SEPLN  \textbf{2150},
  214--228 (2018)

\bibitem{fortuna-etal-2019-hierarchically}
Fortuna, P., Rocha~da Silva, J., Soler-Company, J., Wanner, L., Nunes, S.: A
  hierarchically-labeled {P}ortuguese hate speech dataset. In: Proceedings of
  the Third Workshop on Abusive Language Online. pp. 94--104. Association for
  Computational Linguistics, Florence, Italy (2019)

\bibitem{grave-etal-2018-learning}
Grave, E., Bojanowski, P., Gupta, P., Joulin, A., Mikolov, T.: Learning word
  vectors for 157 languages. In: Proceedings of the Eleventh International
  Conference on Language Resources and Evaluation ({LREC} 2018). European
  Language Resources Association (ELRA), Miyazaki, Japan (2018)

\bibitem{ibrohim-budi-2019-multi}
Ibrohim, M.O., Budi, I.: Multi-label hate speech and abusive language detection
  in {I}ndonesian {T}witter. In: Proceedings of the Third Workshop on Abusive
  Language Online. Association for Computational Linguistics, Florence, Italy
  (2019)

\bibitem{isaksen-gamback-2020-using}
Isaksen, V., Gamb{\"a}ck, B.: Using transfer-based language models to detect
  hateful and offensive language online. In: Proceedings of the Fourth Workshop
  on Online Abuse and Harms. Association for Computational Linguistics, Online
  (Nov 2020)

\bibitem{kang2020hate}
Kang, M., Hall, P.: Hate Speech in Asia and Europe: Beyond Hate and Fear.
  Routledge Contemporary Asia, Taylor \& Francis Group (2020)

\bibitem{kim-2014-convolutional}
Kim, Y.: Convolutional neural networks for sentence classification. In:
  Proceedings of the 2014 Conference on Empirical Methods in Natural Language
  Processing ({EMNLP}). Association for Computational Linguistics, Doha, Qatar
  (2014)

\bibitem{726791}
{Lecun}, Y., {Bottou}, L., {Bengio}, Y., {Haffner}, P.: Gradient-based learning
  applied to document recognition. Proceedings of the IEEE  \textbf{86}(11),
  2278--2324 (1998)

\bibitem{NhaNguyen2020Gender}
Nguyen, T.N., McDonald, M., Nguyen, T.H.T., McCauley, B.: Gender relations and
  social media: a grounded theory inquiry of young vietnamese women’s
  self-presentations on facebook. Gender, Technology and Development pp. 1--20
  (2020)

\bibitem{sanh2020distilbert}
Sanh, V., Debut, L., Chaumond, J., Wolf, T.: Distilbert, a distilled version of
  bert: smaller, faster, cheaper and lighter (2020)

\bibitem{schmidt2017survey}
Schmidt, A., Wiegand, M.: A survey on hate speech detection using natural
  language processing. In: Proceedings of the Fifth International workshop on
  natural language processing for social media. pp. 1--10 (2017)

\bibitem{tuckwood2017hatebase}
Tuckwood, C.: Hatebase: Online database of hate speech. The Sentinal Project.
  Available at: https://www. hatebase. org  (2017)

\bibitem{vaswani2017attention}
Vaswani, A., Shazeer, N., Parmar, N., Uszkoreit, J., Jones, L., Gomez, A.N.,
  Kaiser, {\L}., Polosukhin, I.: Attention is all you need. In: Advances in
  neural information processing systems. pp. 5998--6008 (2017)

\bibitem{sonvx2019}
Vu, X.S., Vu, T., Tran, M.V., Le-Cong, T., Nguyen, H.T.M.: {HSD} shared task in
  {VLSP} campaign 2019: Hate speech detection for social good. In: Proceedings
  of VLSP 2019 (2019)

\bibitem{waseem-hovy-2016-hateful}
Waseem, Z., Hovy, D.: Hateful symbols or hateful people? predictive features
  for hate speech detection on {T}witter. In: Proceedings of the {NAACL}
  Student Research Workshop. pp. 88--93. Association for Computational
  Linguistics, San Diego, California (2016)

\bibitem{yang2019comprehensive}
Yang, X., Yang, L., Bi, R., Lin, H.: A comprehensive verification of
  transformer in text classification. In: China National Conference on Chinese
  Computational Linguistics. pp. 207--218. Springer (2019)

\bibitem{zampieri-etal-2019-predicting}
Zampieri, M., Malmasi, S., Nakov, P., Rosenthal, S., Farra, N., Kumar, R.:
  Predicting the type and target of offensive posts in social media. In:
  Proceedings of the 2019 Conference of the North {A}merican Chapter of the
  Association for Computational Linguistics: Human Language Technologies,
  Volume 1 (Long and Short Papers). Minneapolis, Minnesota (2019)

\end{thebibliography}

\end{document}